\documentclass{article}

\usepackage[final]{neurips_2026}

\usepackage[utf8]{inputenc}
\usepackage[T1]{fontenc}
\usepackage{hyperref}
\usepackage{url}
\usepackage{booktabs}
\usepackage{amsfonts}
\usepackage{nicefrac}
\usepackage{microtype}
\usepackage{xcolor}
\usepackage{graphicx}
\usepackage{subcaption}
\usepackage{algorithmic}
\usepackage{algorithm}
\usepackage{bbm}
\usepackage{enumerate}
\usepackage{caption}
\usepackage{wrapfig}
\usepackage{tabularx,ragged2e}
\usepackage{enumitem}
\usepackage[most]{tcolorbox}
\usepackage{inconsolata}
\usepackage{upquote}
\usepackage{multirow}

\newtcolorbox{figuredbox}{
    enhanced,
    boxrule=0.5pt,
    colback=gray!10,
    colframe=black,
    boxsep=2mm,
    left=1mm,
    right=1mm,
    top=1mm,
    bottom=1mm,
    breakable,
    fontupper=\sffamily\small,
    fonttitle=\sffamily\small,
    width=\textwidth,
    valign=center,
}

\usepackage{amsmath}
\usepackage{amssymb}
\usepackage{mathtools}
\usepackage{amsthm}

\theoremstyle{plain}
\newtheorem{theorem}{Theorem}[section]

\theoremstyle{definition}
\newtheorem{definition}[theorem]{Definition}
\newtheorem{assumption}[theorem]{Assumption}
\theoremstyle{remark}

\usepackage[textsize=tiny]{todonotes}

\colorlet{revisioncolor}{red}

\DeclareCaptionFont{revisioncolor}{\color{revisioncolor}}

\begin{document}
\title{Natural Language Actor-Critic: \\
  \Large{Scalable Off-Policy Learning in Language Space}}

\author{%
  Joey Hong\\
  University of California, Berkeley\\
  \texttt{joey\_hong@berkeley.edu}
  \And
  Kang Liu\\
  ByteDance Seed
  \And
  Zhan Ling\\
  ByteDance Seed
  \And
  Jiecao Chen\\
  ByteDance Seed
  \And
  Sergey Levine\\
  University of California, Berkeley
}

\maketitle

\begin{abstract}
Large language model (LLM) agents---LLMs that dynamically interact with an environment over long horizons---have become an increasingly important area of research, enabling automation in complex tasks involving tool-use, web browsing, and dialogue with people. 
In these settings, traditional policy gradient methods may suffer from unstable learning and poor sample-complexity due to poor credit assignment.
Meanwhile, actor-critic methods address long horizons by learning a critic that provides more granular feedback and enables off-policy (and potentially offline) learning, but are heavily dependent on the fidelity of the critic.
In this paper, we propose \emph{Natural Language Actor-Critic} (NLAC), a novel actor-critic algorithm that trains LLM policies using a generative LLM critic that produces values in natural language space.
While natural language values have been leveraged in the past to provide a more flexible and actionable training signal, our work is the first to perform general and scalable value learning without relying on in-context aggregation of information from on-policy rollouts. 
This means our approach can be trained off-policy without policy gradients, offering a more sample-efficient alternative to existing methods.
We present results on a mixture of reasoning, web browsing, and tool-use with dialogue tasks, demonstrating that NLAC shows promise in outperforming existing training approaches and offers a more scalable and stable training paradigm for LLM agents.
\end{abstract}

\section{Introduction}
While LLMs excel at natural language tasks like question-answering \citep{pyatkin2022reinforced} and problem-solving \citep{hendrycksmath2021,jimenez2024swebench}, which can be solved with a single response, LLM agent tasks require multi-turn interactions. 
Such tasks include autonomous reasoning \citep{deepresearch}, tool-use \citep{nakano2022webgpt}, and dialogue with users \citep{hong2023zeroshot,yu2023promptbasedmontecarlotreesearch}.
 These tasks require agents to dynamically plan and respond to environmental stimuli, which base LLMs struggle to do without additional fine-tuning \citep{bachmann2024pitfalls}. 
To train effective LLM agents, we will need algorithms that can fine-tune LLMs to pursue temporally extended goals within multi-turn interactions.
However, in long-horizon settings where sparse rewards offer poor credit assignment, training LLM policies naively can be unstable and inefficient.

Actor-critic methods tackle long-horizon tasks by learning a critic that evaluates actions and provides a denser training signal, improving sample efficiency and enabling off-policy or offline learning.
However, such methods are highly dependent on the quality of the critic, and prior attempts at actor-critic methods for LLM policies have not demonstrated convincing improvement over standard fine-tuning techniques~\citep{sodhi2023effectiveness,abdulhai2023lmrl}.
We posit that this is because standard scalar-valued critics are a poor fit for
language action spaces: they compress rich failure modes
into a single scalar and provide little guidance on how a language policy should improve.

To address the drawbacks of standard critics for language policies, \citet{feng2025naturallanguagereinforcementlearning} propose Natural Language Reinforcement Learning (NLRL), which frames policy evaluation and improvement directly in language space. 
Natural language values can provide richer, more flexible credit assignment than scalar critics, since they can explain \emph{why} an action is good or bad and suggest how it should be refined.
However, existing approaches to NLRL typically rely on aggregating information from many on-policy rollouts in-context. This becomes increasingly unscalable as trajectories grow longer and environment dynamics become more complex; hence, thus far NLRL has largely been evaluated in simple environments such as gridworld navigation.

In this paper, we propose Natural Language Actor-Critic (NLAC), a novel algorithm for training LLM agents, where a natural language critic is jointly trained with a policy in a general and scalable manner.
The key insight that allows NLAC to address the limitations of NLRL is the training of the critic to predict infinite-horizon futures as \emph{compact summaries}. This is done via a novel \emph{language Bellman backup} objective that bootstraps from one-step transitions, and completely circumvents in-context aggregation over on-policy samples and enables scalable off-policy value learning in language space.
Theoretically, we show that under explicit representation-realizability and order-preservation assumptions, the fixed point of our critic objective encodes successor features~\citep{barreto2018successorfeaturestransferreinforcement} and preserves the ordering of the true scalar Q-values. 
Empirically, we evaluate our approach on a range of LLM agent tasks, ranging from reasoning, tool-use, and dialogue.
Our empirical results demonstrate substantial improvement over prior approaches to learn LLM agents, showing our algorithm is an appealing alternative to prevailing on-policy training methods.

\vspace{-0.1in}
\section{Related Work}
\vspace{-0.1in}
\textbf{LLM agents.}
LLM agents can be used to tackle a variety of complex real-world tasks, including dialogue~\citep{hong2023zeroshot,yu2023promptbasedmontecarlotreesearch}, tool-use~\citep{nakano2022webgpt,schick2023toolformerlanguagemodelsteach}, and embodied decision-making ~\citep{wang2023voyager}. 
The primary challenge in the design of effective LLM agents is enabling LLMs, which traditionally excel at generating single-step responses, to interact sequentially with an environment to accomplish a long-term objective. 
ReAct prompting is a popular method to leverage chain-of-thought reasoning of LLMs for long-horizon planning, by instructing LLMs to explicitly articulate their high-level plans \citep{yao2022react}.
More recent approaches have explored the capability of LLM agents to self-correct their initial attempts at planning using more sophisticated prompting techniques ~\citep{shinn2023reflexionlanguageagentsverbal,madaan2023selfrefineiterativerefinementselffeedback,zhou2024languageagenttreesearch}. 
For example, Reflexion prompting adds a step of self-reflection on top of ReAct to allow LLM agents to refine their initial reasoning after some environment feedback ~\citep{shinn2023reflexionlanguageagentsverbal}. 
However, self-correction methods rely the ability to ``backtrack," or undo previous actions, whereas we measure the capability of LLM agents with one chance to solve a task. 

\textbf{Process reward models.} 
One of the primary challenges in learning LLM agents is the reliance on a single, sparse reward for long-horizon interactions. This makes credit assignment, or distinguishing between good and bad actions in a long rollout, difficult. Process reward models (PRMs) aim to address this, particularly by providing action-level feedback using either human annotations ~\citep{lightman2023letsverifystepstep}, or an estimated value function in the absence of human intervention~\citep{wang2024mathshepherdverifyreinforcellms,setlur2024rewardingprogressscalingautomated}. Our learned natural language critic can be considered an instance of an PRM, but unlike traditional PRMs that provide scalar feedback over actions, our critic outputs feedback in language space. We believe such feedback is more useful for LLM policies that can understand and articulate their decisions in natural language.

\textbf{Reinforcement learning for LLM agents.}
More recently, multiple works have attempted to explicitly fine-tune LLMs as agents using RL \citep{carta2024groundinglargelanguagemodels,zhou2024archertraininglanguagemodel}. The primary way this was done was naively adapting traditional RL fine-tuning used to align LLM responses to multi-turn tasks with environment interaction \citep{stiennon2020learning,ouyang2022training,ramamurthy2023is}. These methods used PPO \citep{schulman2017proximal} to finetune LLMs using the environment reward.  
However, traditional policy optimization for long-horizon tasks exacerbates the instabilities of RL training, particularly due to reliance on exploration and proper credit assignment. 
In this work, we hypothesize that training in natural language over scalar space improves stability and sample efficiency.

\textbf{RL from textual feedback.}
The closest work to ours that does this is NLRL \citep{feng2025naturallanguagereinforcementlearning}, which also proposes learning value functions that output text. However, in NLRL, these values are obtained via repeated sampling of on-policy trajectories and aggregating them in-context. 
In addition, policy improvement is achieved by enumerating over possible actions and their evaluations.
We believe such enumeration and aggregation in-context is intractable for tasks with complex dynamics and large action spaces.
Our method circumvents these drawbacks by training the critic to probabilistically generate compact summaries of rollouts via a novel language Bellman backup, then updating a policy through distillation from a refinement policy that conditions on the critic output.
Recently, self-distillation methods have gained popularity as a sample-efficient method to learn from textual critiques~\citep{hubotter2026sdpo,zhao2026self}. However, such methods typically require ground-truth privileged information and work best in single-turn settings such as mathematical reasoning.

\vspace{-0.1in}
\section{Preliminaries}
\vspace{-0.1in}
\label{sec:preliminaries}

\textbf{Markov decision processes.} 
We adopt the formalism of a Markov decision process (MDP) given by $M = (\mathcal{S}, \mathcal{A}, P, r, \rho, \gamma)$, where $\mathcal{S}$ is the state space, $\mathcal{A}$ is the action space, $P$ is the transition function, $r$ is the reward function, $\rho$ is the initial state distribution, and $\gamma$ is the discount factor. When action $a \in \mathcal{A}$ is executed at state $s \in \mathcal{S}$, the next state is sampled $s' \sim P(\cdot | s, a)$, and the agent receives reward $r(s, a) \in \mathbb{R}$. 

\textbf{LLM agents in MDPs.}
Tasks considered by LLM agents can be defined under the MDP formalism as follows. 
Here, the state and action space are finite-length sequences of tokens in vocabulary $\mathcal{V}$, or $\mathcal{S}, \mathcal{A} \subseteq \mathcal{V}^*$, where $\mathcal{V}^*$ denotes all finite sequences comprised of tokens in vocabulary $\mathcal{V}$.
We also define the space of environment observations $\mathcal{O} \subset \mathcal{V}^*$;
those could consist of results of API calls in tool-use applications, or responses by other interlocutors in dialogue. 
The agent corresponds to a policy $\pi$ that starts by observing a task description along with any initial observations $s_1 = (q, o_0)$. At timestep $t$, the agent \emph{state} $s_t$ of the MDP consists of the history of interaction thus far $s_t = (q, a_1, o_1, \hdots, a_{t-1}, o_t)$ consisting of agent actions and environment observations. 

\textbf{ReAct prompting.} LLM agents are commonly implemented using ReAct prompting  to better leverage the base reasoning capabilities of LLMs ~\citep{yao2022react}.
ReAct prompting instructs LLM agents to output actions $a_t \sim \pi(\cdot | s_t)$ that are actually composite, consisting of a \emph{thought} $a_t^\mathsf{tht}$ where the agent performs a reasoning step, followed by the actual environment action $a_t^{\mathsf{env}}$.
For example, in dialogue, the thought could be the high-level strategy or plan the agent aims to execute, whereas the environment action is the actual utterance by the agent.
Then, the transition function appends to $s_t$ the environment action $a_t^{\mathsf{env}}$ as well as any new observations by the environment $o_{t+1}$, to form the next state $s_{t+1}$. 
Note that the thought does not affect the transition dynamics.

\textbf{Reinforcement learning.}
The objective of RL is to find a policy $\pi$ that maximizes the expected discounted return 
$J(\pi) = \mathbb{E}_{\tau \sim p^\pi}\left[\sum_{t = 0}^{T-1}\gamma^t r(s_t, a_t)\right]$ in an MDP, where $\tau = (s_0, a_0, s_1, a_1, \hdots, s_T)$ and $p^\pi(\tau) = \rho(s_0)\prod_{t=0}^{T-1} \pi(a_t | s_t) P(s_{t+1} | s_t, a_t)$. Standard policy gradient approaches directly train policy $\pi$ using the gradient of $\nabla_\pi J(\pi)$, while more sophisticated algorithms such as PPO and GRPO additionally clip the updates to improve stability ~\citep{schulman2017proximal,shao2024deepseekmathpushinglimitsmathematical}. Actor-critic algorithms additionally learn a state-action value function, or Q-function, defined as $Q^\pi(s_t, a_t) = \mathbb{E}_{(s, a)_{t+1:\infty} \sim p^\pi}\left[\sum_{t' = t}^{T-1}\gamma^{t' - t} r(s_{t'}, a_{t'})\right]$. Such Q-functions are learned by regressing to their Bellman backup:
$
\mathcal{B}Q^\pi(s_t, a_t)=r(s_t,a_t)+\mathbb{E}_{s_{t+1},a_{t+1}\sim P^\pi}\!\left[Q^\pi(s_{t+1},a_{t+1})\right]
$,
where $P^\pi(s', a' | s, a) = P(s' | s, a) \pi(a' | s')$.
Then, an improved policy $\pi'$ can be derived using the Q-function via greedy or maximum-entropy optimization.

\textbf{Successor features and distributional RL.} Two classical ideas inspire our critic. Successor features~\citep{barreto2018successorfeaturestransferreinforcement} represent a state-action pair by the expected discounted sum of future state features $\Psi^\pi(s, a)$, so that $Q^\pi(s, a) = \Psi^\pi(s, a) \cdot w$ whenever the reward is linear, $r(s, a) = \phi(s) \cdot w$; distributional RL~\citep{bellemare2017distributional} learns the full distribution of returns rather than its expectation. Both are learned by bootstrapping from one-step transitions, as is our critic, which generates samples from a distribution over textual descriptions of the future.

\begin{figure*}[t]
\centering
\includegraphics[width=\linewidth]{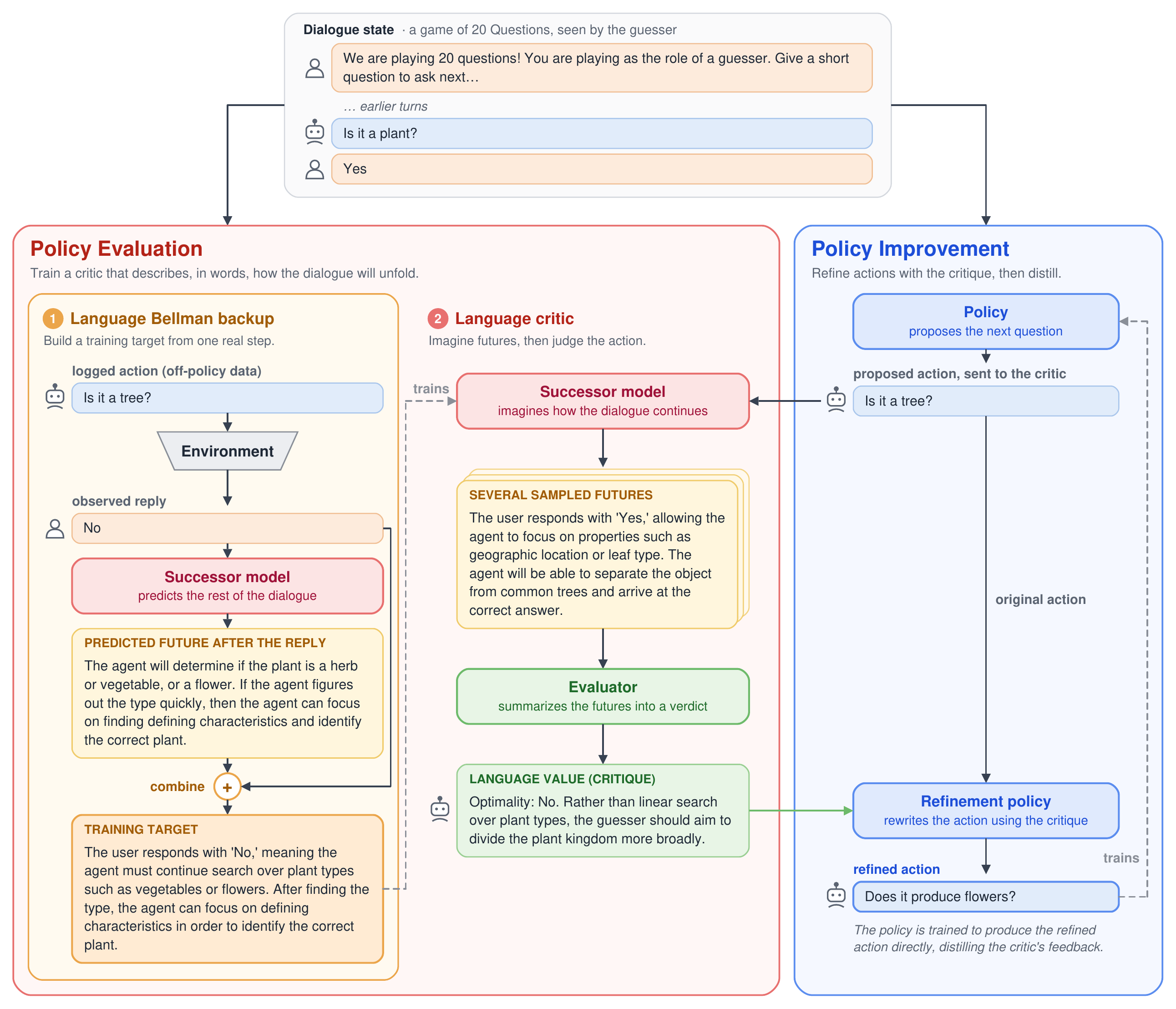}
\caption{\footnotesize Overview of NLAC. During \emph{policy evaluation}, the critic is trained using a language Bellman backup that operates in textual space. During \emph{policy improvement}, the policy is distilled from a refinement policy.}
\label{fig:method}
\vspace{-0.25in}
\end{figure*} 

\vspace{-0.1in}
\section{Natural Language Actor-Critic}
\vspace{-0.1in}

In this section, we present Natural Language Actor-Critic (NLAC), our new method for training LLM agents that adopts the actor-critic paradigm. 
Unlike traditional methods that rely on simple policy gradients, NLAC leverages a \emph{natural language critic} that outputs textual critiques of actions to provide a rich, interpretable, and more stable training signal. Our framework is inspired by classical actor-critic methods where each step consists of (1) policy evaluation, where a critic is trained to assess actions by a policy, and (2) policy improvement, where the policy is updated using evaluations by the critic, but is adapted to leverage the implicit reasoning capabilities of LLMs over text space. 
In our approach, both the LLM policy and the natural language critic are instantiated by the same underlying LLM, with their distinct functionalities realized through different prompts. We go over both components in detail below.

\vspace{-0.1in}
\subsection{Policy Evaluation}
\vspace{-0.1in}

In traditional actor-critic approaches, a critic is trained to estimate scalar state-action values, or Q-values, typically denoted as $Q^\pi(s,a) \in \mathbb{R}$, which represents the expected return by policy $\pi$ from state $s$ after taking action $a$. 
While learning such Q-values can be similarly done with LLM critics, LLMs are better suited to process and generate natural language over scalars. Therefore, we believe evaluation that is in natural language space leverages prior text-based reasoning capabilities of LLMs, and thus will largely improve sample efficiency. 
Hence, our natural language critic is an LLM that generates textual critiques, denoted as $Q^\pi_L(s, a) \in \mathcal{V}^*$, that not only comments on how good an action is, but also explains why.

\textbf{Predicting the future using language.} The key addition that is not captured by scalar Q-values is an explanation of why a particular action is optimal or not. As we will discuss later, this information is valuable for LLM policies to understand how to refine their actions during policy improvement, avoiding the reliance on random exploration to discover better actions. 
We believe that the key for a critic to derive these explanations is the prediction and analysis of future outcomes. 
However, full trajectories can be prohibitively complex to model, so we instead predict compact summaries of futures.
In order to do so, we must train our natural language critic to additionally act as a successor function, defined as follows:

\begin{definition}
A \textbf{language successor model} $M^\pi$ for policy $\pi$ takes a state $s_t$ and action $a_t$ as input, and probabilistically generates a textual summary of rollout $(s, a)_{t+1: T}$, or what will happen to policy $\pi$ in the future, and reward $r(s_T)$. We denote by $M^\pi(\cdot \mid s_t, a_t)$ the distribution from which such descriptions are sampled.
\vspace{-0.1in}
\end{definition}
Our language successor model shares similarities with successor features~\citep{barreto2018successorfeaturestransferreinforcement} in that both can predict a distribution over future rollouts, and---as we show later---be trained using temporal difference learning. The main difference lies in that traditional successor features are used to compute Q-values via a linear product, whereas ours is used to generate state-action values in natural language via output by an LLM.

To train our language successor model, we draw inspiration from distributional value learning~\citep{bellemare2017distributional}, which introduces a distributional Bellman backup to train a distribution over returns rather than just their scalar expectation.
Notably, the distributional Bellman backup used one-step samples of the future and thus could be computed off-policy. 
Similarly, we propose a \emph{language Bellman backup} $\mathcal{B}_L$ that bears some semblance to the distributional Bellman backup, but makes key adaptations to account for samples that are textual descriptions of rollouts.
\begin{definition}
A \textbf{language Bellman backup} $\mathcal{B}_L$ takes a language successor model $M^\pi$, along with state $s_t$ and action $a_t$ as input, and computes distribution $\mathcal{B}_L\,M^\pi(\cdot | s_t, a_t)$ such that the probability of description $d_t \in \mathcal{V}^*$ is given by:
\begin{align}
\mathcal{B}_L\,M^\pi(d_t \mid s_t, a_t)
&= \mathsf{Pr}\Bigl[
    d_t = B\bigl(r(s_t, a_t), s_{t+1}, a_{t+1}, d_{t+1}\bigr)
    \ \Big|\ \ s_{t+1}, a_{t+1}, d_{t+1}
\Bigr] \, , \nonumber \\
&\quad s_{t+1}, a_{t+1} \sim P^\pi(\cdot \mid s_t, a_t), \quad
 d_{t+1} \sim M^\pi(\cdot \mid s_{t+1}, a_{t+1}) \, .
\end{align}
where $B$ is a function that combines immediate next state and action $s_{t+1}, a_{t+1}$ with description $d_{t+1}$ of rollout $(s, a)_{t+2: T}$ into one concatenated description of $(s, a)_{t+1: T}$.
\vspace{-0.1in}
\end{definition}
Beyond simple concatenation, the $B$ function ``discounts" the future rollout description from $M^\pi$ in the concatenated rollout so the immediate next state is given more emphasis in the description.

Then, we can train our language successor model $M^\pi$ by minimizing the divergence between distributions $M^\pi(\cdot | s_t, a_t)$ and target distributions created by the language Bellman backup: $M^\pi = $
\begin{equation}
M^\pi=\arg\min_M\,\mathbb{E}_{(s_t,a_t,s_{t+1})\sim\mathcal{D}}\!\left[D_f\bigl(M(\cdot\mid s_t,a_t)\,\|\,\mathcal{B}_L M(\cdot\mid s_t,a_t)\bigr)\right].
\label{eq:successor_train}
\end{equation}
Note that our training objective is an instance of temporal-difference learning and thus does not require on-policy Monte Carlo trajectories.

\textbf{Generating critiques.} Finally, the natural language critic should analyze all possible futures in order to evaluate how good an action is in expectation, then explain its reasoning by referencing possible future outcomes. 
To perform this evaluation, we define the following:
\begin{definition}
A \textbf{language evaluator} $E$ takes as input state $s_t$ and action $a_t$, along with a sequence of descriptions of possible rollouts $(s, a)_{t+1: T}$ and their rewards $r(s_T)$, and outputs a textual critique that comments on whether $a_t$ was optimal, with justification using possible future outcomes. 
\vspace{-0.1in}
\end{definition}
Then, we can approximate $Q^\pi_L(s_t, a_t)$ as:
\begin{equation}
Q_L^\pi(s_t,a_t)\approx E\bigl(s_t,a_t,d_t^{(1)},\ldots,d_t^{(k)}\bigr),\qquad d_t^{(i)}\sim M^\pi(\cdot\mid s_t,a_t)\ \forall i\in[k].
\label{eq:nlq}
\end{equation}
Note that $E$ essentially aggregates and summarizes multiple descriptions of different rollouts that are all fit in-context, which LLMs have demonstrated a priori efficacy in without additional training~\citep{feng2025naturallanguagereinforcementlearning}. This means that the only training required to perform evaluation of policy $\pi$ in language space is learning the language successor model $M^\pi$ (see Figure~\ref{fig:method} for illustration).


\vspace{-0.1in}
\subsection{Policy Improvement}
\vspace{-0.1in}

Thus far, we showed how to train the natural language critic to evaluate a fixed policy $\pi$. We now show how an improved policy can be learned using
textual critiques $Q_L^\pi(s, a)$ obtained by a critic using Equation~\ref{eq:nlq}. Naturally, such policy is a greedy policy where $a \sim \pi(\cdot|s)$ satisfies $a = \arg\max_{a'} Q^\pi_L(s, a')$. Note that we assume:
\begin{assumption}
For any policy $\pi$, the set $\{Q_L^\pi(s, a')\}_{a' \in \mathcal{A}}$ computed using Equation~\ref{eq:nlq} for any state $s$ forms a totally-ordered set with binary relation $\geq$.
\end{assumption}
We believe that this is not a strong assumption, as each critique $Q^\pi_L(s, a)$ can be mapped to a scalar that quantifies its sentiment, which can be used to compare with other critiques. Then, $Q^\pi_L(s, a') \geq Q^\pi_L(s, a)$ if the underlying sentiment of the text in $Q^\pi_L(s, a')$ is more positive.

However, computing the greedy policy is intractable for LLM agent tasks, where the action spaces $\mathcal{A} \subseteq \mathcal{V}^*$ are combinatorial in the token vocabulary, making it impossible to enumerate all possible actions to find the optimal one. While prior works have proposed sampling a subset of actions and reweighting \citep{li2024QProbe}, we find empirically that for tractable sample sizes, this approach does not sufficiently explore the space of possible actions.

Our approach sidesteps this issue by leveraging the descriptive power of the natural language values using a self-refinement paradigm.
Our insight is that the natural language value $Q^\pi_L(s,a)$ not only comments on how good an action is, but also contains intuition on how a suboptimal action can be improved. Hence, a policy that is an LLM with strong base reasoning capabilities can process this evaluation and understand how to \emph{refine} its initial action.  

To this end, we define a \emph{refinement policy} $\pi^r$ that takes an action $a_t \sim \pi(\cdot|s_t)$ by the base policy, and generates a refined action $a^{r}_t \sim \pi^r(\cdot|s_t, a_t, Q^\pi_L(s_t, a_t))$ that is better according to the natural language critic, i.e., $Q^\pi_L(s_t, a_t^{r}) \ge Q^\pi_L(s_t, a_t)$. 
As with the policy and critic, the refinement policy can use the same underlying LLM but with a different prompt. Note that refinement can also be performed iteratively by maintaining and appending to a history of all previous action attempts and their evaluations 
\begin{equation}
a_t^{r}\sim\pi^r\!\left(\cdot\mid s_t,a_t^{1},Q_L^\pi(s_t,a_t^{1}),\ldots,Q_L^\pi(s_t,a_t^{m})\right).
\end{equation}
where we can control for a parameter $m$ that denotes number of rounds of refinement.
As $m \to \infty$, we expect the refined action $a_t^{r}$ to be the greedily optimal one $a_t^{r} = \arg\max_a Q^\pi_L(s_t, a)$.

Finally, we propose a policy improvement objective from $\pi$ to $\pi'$ that projects the refinement policy back to the base policy, similar to the policy updates in SAC \citep{haarnoja2018soft}.
However, rather than parameterizing a target policy using the learned values, which requires enumeration over actions and is intractable in our setting, we let the target policy be the refinement policy:
\begin{equation}
\pi'=\arg\max_\pi\,\mathbb{E}_{s_t\sim\mathcal{D}}\!\left[D_f\bigl(\pi(\cdot\mid s_t)\,\|\,\pi^r(\cdot\mid\ldots,Q_L^\pi(s_t,a_t^m))\bigr)\right].
\label{eq:policy_train}
\end{equation}
In practice, we found that a single round of refinement $m=1$ was sufficient. Again, this objective does not require any on-policy rollouts, and can therefore be trained off-policy.

\textbf{Why distill from a refinement policy?} Our policy update is supervised fine-tuning on critique-conditioned refined actions rather than a policy-gradient update. This still follows the actor-critic template of alternating policy evaluation and improvement: like SAC~\citep{haarnoja2018soft}, we improve the policy by projecting it onto a better target policy via KL minimization. Conditioning on the critique also aids exploration, as the refinement policy can propose actions that are unlikely under the base policy (see examples in Appendix~\ref{app:examples}).

\vspace{-0.1in}
\subsection{Theoretical Results}
\vspace{-0.1in}
\label{sec:analysis}

Due to space, we provide a theoretical analysis of NLAC in Appendix~\ref{app:proofs}. At a high level, our analysis connects the representations used by our learned critic to successor features, which allows us to recover standard scalar value-learning guarantees even though the critic outputs text. 

Our main results state that, under realizability assumptions on the reward and on how the language Bellman backup composes representations, we show that the learned language critic can be used to recover the true Q-function:
\begin{theorem}
Consider policy evaluation via Equation~\ref{eq:successor_train} and let $Q_L^\pi$ be the natural language critic at convergence. For any state $s$ and action $a$, there exists monotonic mapping $g$ such that $Q^\pi(s, a) = g(Q_L^\pi(s, a))$, where $Q^\pi$ denotes the true scalar Q-function.
\end{theorem}

While it is true that such assumptions are difficult to ensure in practice, the main contribution of this analysis is connecting prediction of future in text space with learning better value functions. Note that convergence to the optimal policy (Theorem~\ref{thm:iteration} in Appendix~\ref{app:proofs}) additionally requires the refinement policy to act as an oracle policy-improvement operator, which is admittedly a strong assumption. The main message of our analysis is thus that, under these assumptions, the fixed point of our critic objective encodes successor features and preserves the ordering of scalar Q-values.

\vspace{-0.1in}
\section{Practical Implementation}
\vspace{-0.1in}
\label{sec:implementation}

In this section, we describe how both the critic and policy are trained in practice. We defer specific details such as exact prompts used to Appendix~\ref{app:imp_details}. Though our method involves many different components such as a language successor model and evaluator, we can leverage the general capabilities of LLMs to reason over and generate language to reuse one model to implement all the described components.
Hence, our algorithm only involves training one LLM $\mathcal{M}$ with parameter $\theta$. For a prompt $p \in \mathcal{V}^*$, we denote by $\mathcal{M}_\theta(p)$ the distribution over responses by the LLM. 

\vspace{-0.1in}
\subsection{Training Components}
\vspace{-0.1in}

\textbf{Policy.} Many prior works have parameterized policies as LLMs. One of the greatest advantages of doing so is the ability to leverage the strong reasoning capabilities of LLMs from chain-of-thought prompting \cite{wei2023chainofthought, yao2022react}. By choosing a proper prompt $p_{\text{react}}$, an LLM policy can be instructed to describe their underlying thoughts for choosing a particular action in addition to generating the action itself $a_t \sim \mathcal{M}_\theta(p_{\text{react}}(s_t))$.  

\textbf{Language successor model.} LLMs have demonstrated efficacy at predicting realistic future rollouts in a variety of environments \citep{lin2024learningmodelworldlanguage}. 
These futures are generated by simply processing the state-action in a prediction prompt $p_{\text{pred}}$ that also instructs the LLM to summarize rollouts into concise textual descriptions, then sampling from the LLM output $M_{\theta}(\cdot \mid s_t, a_t) = \mathcal{M}_\theta(p_{\text{pred}}(s_t, a_t))$.

\textbf{Language Bellman backup.} The backup $\mathcal{B}_L$ also outputs a distribution over descriptions of rollouts, but uses one-step samples of next state along with a ``bootstrapped" description of rollout generated by $M_\theta$. 
We give the LLM instruction $p_{\text{tpred}}$ to predict a ``target" future by combining the immediate next state with the bootstrapped future description into one description, discounting the future description as necessary by placing more emphasis on the immediate next state.
\begin{align*}
\qquad \mathcal{B}_L M_\theta(\cdot \mid s_t, a_t)
= \mathcal{M}_\theta\bigl(p_{\text{tpred}}(r_t, s_{t+1}, d_{t+1})\bigr)\,,\ \text{ where } d_{t+1} \sim \mathcal{M}_{\theta}\bigl(p_{\text{pred}}(s_{t+1})\bigr)\,.
\end{align*}
Note that we do not explicitly sample $a_{t+1}$ from the policy, but implicitly via the language successor model that is conditioned on the policy. 

\textbf{Language evaluator.} The evaluations by $E$, which ultimately become the outputs of the natural language critic that estimate $Q_L^\theta(s_t, a_t)$ can similarly be derived by fitting multiple generated futures $d_t^{(1)}, \hdots d_t^{(k)}$ in-context within an evaluation prompt $p_{\text{eval}}$ that asks the LLM to aggregate the futures and summarize into an overall evaluation of an action as 
$Q_L^\theta(s_t, a_t) \sim \mathcal{M}_\theta(p_{\text{eval}}(d_t^{(1)}, \hdots, d_t^{(k)}))$. 

\textbf{Refinement policy.} Finally, the refinement policy $\pi^r$ can also be obtained by an LLM instructed to refine its latest action given an evaluation similar to prior self-refinement approaches \citep{madaan2023selfrefineiterativerefinementselffeedback}. The refined action is obtained via prompt $p_{\text{refine}}$ as 
$
a_t^r\sim\mathcal{M}_\theta\!\left(p_{\mathrm{refine}}(s_t,a_t^{1},\ldots,Q_L^\theta(s_t,a_t^{m}))\right)
$.

\vspace{-0.1in}
\subsection{Training Algorithm}
\vspace{-0.1in}

Formally, the parameters $\theta$ are trained using two objectives for policy evaluation and improvement. For policy evaluation, for a transition $(s_t, a_t, s_{t+1})$, the natural language critic is trained using cross entropy component of the objective:
\begin{equation*}
\mathcal{L}_1(s_t,a_t,s_{t+1})=D_{\mathsf{KL}}\!\left(\mathcal{B}_L M_{\bar{\theta}}(\cdot\mid s_t,a_t)\,\|\,M_\theta(\cdot\mid s_t,a_t)\right).
\end{equation*}
where $\bar{\theta}$ are reference parameters that are an exponentially moving average of the trained parameters, in order to prevent generative collapse ~\citep{Shumailov2024AIMC}. 
We choose the reverse direction of KL-divergence to capture the full diversity over possible futures. 
Then, for policy improvement, we train the policy on the log-likelihood loss:
\begin{equation*}
\qquad\qquad \mathcal{L}_2(s_t,k)=-\log\pi_\theta(a_t^r\mid s_t),\qquad a_t\sim\pi_\theta(\cdot\mid s_t),\ \ a_t^r\sim\pi^r_\theta(\cdot\mid\ldots,Q_L^\theta(s_t,a_t)).
\end{equation*}
This objective can be interpreted as distillation, but using generations by the refinement policy as the teacher policy. Note that the loss depends on $k$ via $Q_L^{\theta}(s_t, a_t)$ given by Equation~\ref{eq:nlq}.
We provide pseudocode for NLAC in Appendix~\ref{app:pseudocode}.
Note that by default, our refinement policy relies on the base reasoning capabilities of the pretrained LLM. In Appendix~\ref{app:refinement_training}, we discuss how the refinement policy can be explicitly trained.
We also discuss effective methods for mitigating catastrophic forgetting to enable training over longer timescales in Appendix~\ref{app:lwf}.



\vspace{-0.1in}
\section{Experiments}
\vspace{-0.1in}

To demonstrate the effectiveness of NLAC, we evaluate our method on a variety of LLM agent tasks:  mathematical reasoning~\citep{hendrycksmath2021}, strategic dialogue ~\citep{20Q}, and customer service using mixed dialogue and tool-use~\citep{yao2024taubenchbenchmarktoolagentuserinteraction}. Though mathematical reasoning does not involve interaction with an environment, it is currently the most popular benchmark to evaluate different RL fine-tuning algorithms. 

\vspace{-0.1in}
\subsection{Task Descriptions}
\vspace{-0.1in}

\textbf{Mathematical reasoning.} 
We evaluate on mathematical problem-solving using the MATH dataset ~\citep{hendrycksmath2021}, which consists of different competition math problems of varying level of difficulty. A score of $1$ is achieved if the agent solves the problem and outputs an answer that is correct and properly formatted. 
We evaluate on a subset of $500$ problems from the test dataset of the highest difficulty level, which we call MATH500-Hard. The remaining $12,000$ problems are used as the training set for RL fine-tuning.

\textbf{Dialogue game.} We use the popular game of 20 Questions (20Q) as a representative strategic dialogue task, where the LLM agent acts as the guesser to uncover the hidden word by an oracle. 20Q was chosen because it was non-adversarial (so we can evaluate against a fixed LLM as the oracle), and requires the LLM agent to generate a cohesive sequence of actions over multiple steps. 
Though many implementations exist ~\citep{srivastava2023beyond,abdulhai2023lmrl}, we follow the one by \citet{20Q} where the set of hidden words can be any in a set of $1,823$ objects from the THINGS dataset~\citep{hebart2019things}. 
A reward of $1$ is achieved if the guesser correctly identifies the hidden object within $20$ turns, or questions, where correctness if determined by using the oracle LLM as a judge. We use GPT5~\citep{singh2025openaigpt5card} as the oracle. 
We construct a training set of $1,000$ objects and a test set of $500$ different objects through random sampling. 

\textbf{Customer service.} We consider $\tau$-bench as a representative LLM agent task that requires a mixture of dialogue and tool-use to solve \citep{yao2024taubenchbenchmarktoolagentuserinteraction}. The LLM agent must act as a customer service representative in various scenarios such as modifying items in an user's order, and follow a rigid set of policy guidelines. At every step, the LLM agent can either communicate with the user, or make an API call that interacts with a backend database. At the end, the agent receives a score of $1$ if the database entries match ground-truth values, and the agent did not violate any policy guidelines via their actions. Users are simulated using a GPT5~\citep{singh2025openaigpt5card} model prompted with both an initial request (such as modifying or cancelling an order) as well an identity that can be verified using the database. There are two categories of scenarios: (1) in retail, the LLM agent must modify pending orders of items, return or exchange delivered orders, or update user information, and (2) in airline, the LLM agent must book, modify, or cancel flight reservations. To test generalization, we compile a training dataset of $2,500$ user scenarios in the retail category, and evaluate on a test set of $500$ different retail scenarios, as well as $500$ airline scenarios. 

\vspace{-0.1in}
\subsection{Results}
\vspace{-0.1in}

\begin{table*}[t]
\centering
\small
\setlength{\tabcolsep}{4pt}
\begin{tabular}{@{}p{2.7cm}p{4.5cm}cccc@{}}
\toprule
 & &  \multicolumn{1}{c}{MATH500-Hard}
 & \multicolumn{1}{c}{20Q}
 & \multicolumn{2}{c}{$\tau$-Bench}
 \tabularnewline \cmidrule(l){5-6} 
Paradigm & Method & Accuracy & Winrate & Retail & Airline \\
\midrule
Prompting GPT5 & ReAct \citep{yao2022react} & $80.8$ & $30.2$ & $0.44$ & $0.32$ \\
\midrule
\multirow{5}{2.7cm}{Fine-tuning \\ Qwen2.5-7B-Instruct} 
  & RFT & $49.8$ & $12.6$ &  $0.21$ & $0.13$ \\
  & PPO ~\citep{schulman2017proximal} &  $52.3$ & $17.2$ &  $0.31$ & $0.14$\\
  & GRPO ~\citep{shao2024deepseekmathpushinglimitsmathematical} & $52.5$ & $18.4$ &  $0.32$ & $0.11$\\
  & Self-Distill ~\citep{hubotter2026sdpo} & $\mathbf{57.1}$ & $10.4$ &  $0.24$ & $0.14$ \\
  & NLRL ~\citep{feng2025naturallanguagereinforcementlearning} &  $53.4$ & $25.8$ & $0.25$ & $0.16$ \\
  & NLQL (ablation)  &  $56.4$ & $24.2$ & $0.29$ & $0.19$ \\ 
  & NLAC (online; ours) &  $56.2$ & $26.0$ & $\mathbf{0.42}$ & $\mathbf{0.22}$ \\
  & NLAC (replay; ours) &  $53.2$ & $\mathbf{26.4}$ & $0.39$ & $0.21$ \\
\midrule
\multirow{5}{2.7cm}{Fine-tuning \\ QwQ-32B} 
  & RFT & $69.4$ & $22.0$ &  $0.35$ & $0.29$ \\
  & PPO ~\citep{schulman2017proximal} & $\mathbf{73.4}$ & $24.0$ &  $0.52$ & $0.41$\\
  & GRPO ~\citep{shao2024deepseekmathpushinglimitsmathematical}  & $71.8$ & $25.6$ &  $0.49$ & $0.39$\\
  & Self-Distill ~\citep{hubotter2026sdpo} & $70.1$ & $26.8$ &  $0.35$ & $0.29$ \\
  & NLRL ~\citep{feng2025naturallanguagereinforcementlearning} & $71.5$ & $31.8$ & $0.44$ & $0.31$ \\
  & NLQL (ablation) & $71.2$ & $31.4$ & $0.51$ & $0.36$ \\ 
  & NLAC (online; ours) & $72.7$ & $32.1$ & $0.56$ & $\mathbf{0.43}$ \\
  & NLAC (replay; ours) & $71.9$ & $30.8$ & $\mathbf{0.58}$ & $0.41$ \\
\bottomrule
\end{tabular}
\caption{\footnotesize Performance on evaluation set of each benchmark. To make comparisons fair, each of the fine-tuning methods train for $30,720$ gradient steps in batches of size $1,024$, and we average performance across three independent runs.
Across the board, NLAC outperforms other RL methods for both small- and medium-sized LLMs, and even beats frontier models on long-horizon tasks. Note that because mathematical reasoning is a single-step task, we have to adapt the language generative model to only predict reward.  
}
\vspace{-0.25in}
\label{tab:results}
\end{table*}

We compare NLAC with $k=1$ and $m=1$ against both prompting and fine-tuning baselines. We found those settings of hyperparameters was sufficient to achieve good performance, though more stochastic environments may warrant higher $k$. For baselines that involve fine-tuning, we consider two LLMs: Qwen2.5-7B-Instruct~\citep{qwen2.5}, and QwQ-32B~\citep{qwq32b}, which is also trained on reasoning traces. We choose these two LLMs to measure the effect of increasing size and pre-training on reasoning traces on the performance of the RL methods. Our baselines can be categorized into the following (training details can be found in Appendix~\ref{app:training_details}):

\textbf{Prompting.} We perform ReAct prompting ~\citep{yao2022react} of a state-of-the-art frontier model GPT5~\citep{openai2024gpt4technicalreport}. Because such models do not expose weights for RL fine-tuning, we rely on the zero-shot capabilities of the LLM without any additional training on the tasks. 

\textbf{Rejection fine-tuning.} We perform RFT where at every iteration, the base LLM policy collects a set of on-policy rollouts, of which successful ones are used to train the LLM via SFT.

\textbf{Traditional RL.} The most standard way to perform RL fine-tuning is to train the LLM to optimize score using a policy gradient algorithm on on-policy rollouts. We consider both PPO ~\citep{schulman2017proximal} and GRPO ~\citep{shao2024deepseekmathpushinglimitsmathematical} as the algorithm, the difference being that PPO additionally learns a token-level value function on Monte-Carlo rollouts as a baseline to stabilize reward, whereas GRPO computes the average reward across $8$ different rollouts. 

\textbf{RL with textual feedback.}
We consider recently proposed algorithms that leverage rich feedback in text space. 
Since we do not assume any ground-truth privileged information, the feedback simply consists of $4$ independent, on-policy rollouts. 
First, we compare against a variation of SDPO~\citep{hubotter2026sdpo} that we call Self-Distill, where the rollouts are directly fed to the refinement policy in-context, similar to as in \citet{shinn2023reflexionlanguageagentsverbal}.
Finally, we also compare against NLRL~\citep{feng2025naturallanguagereinforcementlearning}. NLRL additionally aggregates these rollouts to generate a summary evaluation as value without any training, then enumerates over multiple actions and distills the action with highest value. Note that these approaches require $5\times$ more rollouts per gradient step.

\textbf{Ablations.} The main algorithmic contributions of NLAC involve learning a language successor function to tractably extract values in language space, then distilling from a refinement policy that conditions on the learned values. Self-Distillation and NLRL already act as ablations that directly sample additional on-policy rollouts instead of learning a successor function. As another ablation, we compare against NLQL (Q-learning), where instead of distilling from a refinement policy, we sample actions and distill the one that achieves the highest value. NLQL can also be considered an instance of NLRL due also employing value learning in language space. Together, these comparisons isolate each component of NLAC: Self-Distill replaces critiques with raw environment feedback, NLRL replaces the trained critic with critiques generated by aggregating rollouts in-context, and NLRL and NLQL replace the refinement policy with sampling several actions from the base policy and distilling the best one.

\begin{wrapfigure}{r}{0.42\columnwidth}
\vspace{-1.0\baselineskip}
\centering
\includegraphics[width=0.40\columnwidth]{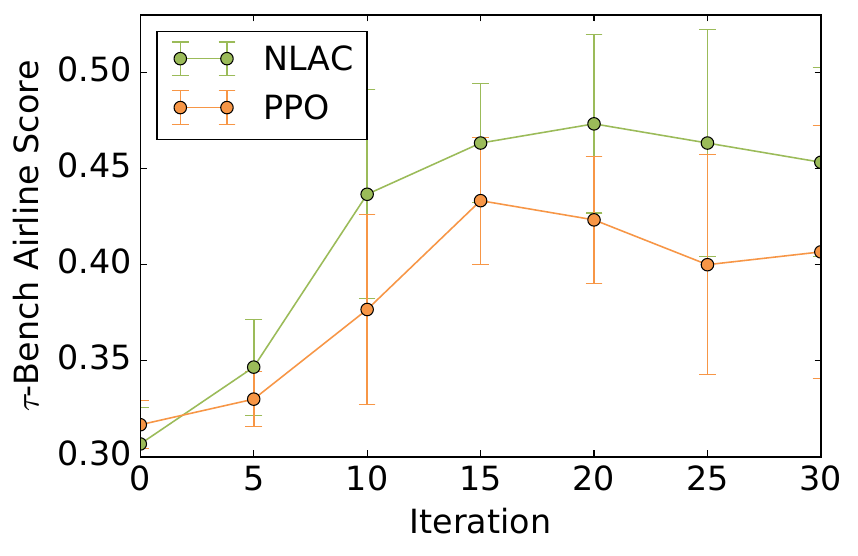}
\caption{\footnotesize Learning curves for NLAC and PPO across three independent runs. NLAC converges in fewer samples.}
\label{fig:learning_curves}
\vspace{-1.0\baselineskip}
\end{wrapfigure}
\textbf{Results.} The results of our evaluation are presented in Table~\ref{tab:results}. We see that for both LLM models, NLAC outperforms other fine-tuning approaches, and even prompting of a significantly stronger LLM, particularly on long-horizon tasks. The only task where NLAC does not outperform other methods is mathematical reasoning, which is a single-step task, in which NLAC reduces to performing self-distillation using a generative reward model~\citep{madaan2023selfrefineiterativerefinementselffeedback}; this is because in single-step tasks, our natural language critic is only trained to predict reward. Though math involves many internal reasoning steps, only one action is taken in the environment, so the critic cannot provide action-level credit assignment, and NLAC behaves like self-refinement methods such as STaR~\citep{zelikman2022star}.
Meanwhile, on tasks requiring multi-step interaction, which our method is tailored for, NLAC greatly outperforms all baselines, achieving a $30\%$ improvement in 20Q and $\tau$-retail over standard RL fine-tuning.
Even when compared against methods that leverage additional on-policy samples, NLAC performed noticeably better, particularly on $\tau$-bench, which is the most difficult task we train on.
Overall, this serves as evidence that NLAC is more suitable to complex tasks. 
Additional qualitative examples demonstrating this are provided in Appendix~\ref{app:examples}.
Furthermore, as shown in Figure~\ref{fig:learning_curves}, NLAC also requires fewer training iterations to achieve maximum performance.

We evaluate two versions of NLAC---\emph{online} and \emph{replay}. In the \emph{online} version, we adopt the typical RLVR setting where on-policy samples are generated for each train-step and not reused. This is the setting all other methods are implemented in. Since our method can also work in the off-policy regime, in the \emph{replay} version, we preserve training samples in a replay buffer and use an equal mix of fresh on-policy samples and stale ones from the buffer for each training batch. This means the \emph{replay} version requires $2\times$ fewer samples than other approaches. The fact that \emph{replay} performs on par with \emph{online} demonstrates a compelling advantage of NLAC, as most RLVR training pipelines are bottlenecked by the generation of training samples. 
This efficiency gain is even further boosted by the fact that many existing RL algorithms reject groups of rollouts containing too much negative signal. Since our method can learn equally well from negative samples, we actually require up to $9\times$ fewer samples than methods as GRPO and PPO for the same number of gradient steps when training on $\tau$-bench, where a vast majority of trajectories initially are failing.

\vspace{-0.1in}
\subsection{Further Analysis}
\vspace{-0.1in}
\label{sec:critic_eval}

\noindent
\begin{minipage}[t]{0.53\textwidth}
\vspace{0pt}
Beyond final task performance, we evaluate the learned critic in isolation. Because there are no ground-truth critiques, we convert each critique into a scalar by asking a separate LLM to predict the probability of success, and measure the explained variance (EV) of observed returns.
On $\tau$-bench with QwQ-32B as the base model, we compare an untrained language critic obtained by prompting, the NLAC critic trained on $1,280$ retail trajectories, and an action-level scalar Q-function trained by Monte-Carlo (MC) regression on rewards from the same trajectories.
\end{minipage}%
\hfill
\begin{minipage}[t]{0.44\textwidth}
\vspace{0pt}
\centering
\small
\setlength{\tabcolsep}{4pt}
\begin{tabular}{@{}lcc@{}}
\toprule
& \multicolumn{2}{c}{$\tau$-Bench EV $\uparrow$}
\tabularnewline \cmidrule(l){2-3}
Value estimator & Retail & Airline \\
\midrule
Scalar critic (MC) & $0.41$ & $0.28$ \\
\midrule
Constant (mean) & $0.00$ & $0.00$ \\
Untrained critic & $0.18$ & $0.14$ \\
NLAC (ours) & $\mathbf{0.33}$ & $\mathbf{0.26}$ \\
\bottomrule
\end{tabular}
\vspace{-0.5\baselineskip}
\captionof{table}{\footnotesize EV of values extracted from each critic on $100$ held-out trajectories.}
\label{tab:critic_eval}
\end{minipage}

\vspace{0.3\baselineskip}
As shown in Table~\ref{tab:critic_eval}, training with our language Bellman backup nearly doubles EV over the untrained critic, both on the retail scenarios it was trained on and on unseen airline scenarios, indicating that textual bootstrapping preserves useful value information.
The scalar critic achieves higher EV, but it is trained to optimize exactly this objective, whereas our critic is trained in text space. The purpose of a textual critic is a richer signal for policy improvement, and the comparison to NLQL in Table~\ref{tab:results} shows that it leads to a better policy.

\vspace{-0.15in}
\section{Discussion}
\vspace{-0.15in}

In this paper, we propose NLAC, a new actor-critic algorithm for training LLM agents under the paradigm of learning values in natural language space. In our work, the natural language values not only comment on the optimality of an action, but also articulate why by predicting and analyzing future outcomes. The key innovation we propose to enable this is a novel \emph{language Bellman backup} that trains language successor function to generate summaries of future rollouts using only one-step samples obtained off-policy. Then, an LLM policy can be improved by refining its own suboptimal actions. 
Empirically, we show that NLAC greatly outperforms other prompting and fine-tuning baselines on long-horizon tasks involving dialogue and tool-use. 
One apparent limitation of our approach is the reliance on the base LLM's ability to both generate plausible futures and refine its own actions through carefully tuned prompts, which can break when tasks become prohibitively complex. 
An important direction for future research is the removal of this dependency, potentially by adding the ability to decompose complex tasks hierarchically into manageable components. While our policy improvement approach can be considered an instance of self-distillation, we explicitly train a critic to provide the necessary textual feedback, rather than relying on ground-truth information.



\bibliography{paper}
\bibliographystyle{plainnat}

\newpage
\appendix
\onecolumn

\section{Pseudocode}
\label{app:pseudocode}

\begin{algorithm}[H]
    \caption{Natural Language Actor-Critic (NLAC)}
    \label{alg:nlac}
    \begin{algorithmic}[1]
    \STATE Initialize $\theta, \bar{\theta}$ from a pretrained model and initialize replay buffer $\mathcal{D}$.
    \FOR{each iteration}
        \FOR{each environment step}
            \STATE Sample action $a_t \sim \pi_\theta(\cdot \mid s_t)$.
            \STATE Step the environment: $s_{t+1} \sim P(\cdot \mid s_t, a_t)$ and observe reward $r_t$.
            \STATE Add $(s_t, a_t, r_t, s_{t+1})$ to replay buffer $\mathcal{D}$.
        \ENDFOR
        \FOR{each sampled transition $(s_t, a_t, r_t, s_{t+1}) \sim \mathcal{D}$}
            \STATE Sample bootstrapped future $d_{t+1} \sim M_{\bar{\theta}}(\cdot \mid s_{t+1}, a_{t+1})$.
            \STATE Form target description distribution
            \STATE \hspace{1em} $\mathcal{B}_L M_{\bar{\theta}}(\cdot \mid s_t, a_t) = \mathcal{M}_{\bar{\theta}}\bigl(p_{\mathrm{tpred}}(r_t, s_{t+1}, d_{t+1})\bigr)$.
            \STATE Compute critic loss
            \STATE \hspace{1em} $\mathcal{L}_1 = D_{\mathsf{KL}}\!\left(\mathcal{B}_L M_{\bar{\theta}}(\cdot\mid s_t,a_t)\,\|\,M_\theta(\cdot\mid s_t,a_t)\right)$.
            \STATE Sample evaluation $Q_L^{\theta}(s_t,a_t)$ from $p_{\mathrm{eval}}$ using $k$ futures.
            \STATE Sample refined action $a_t^r \sim \pi_\theta^r(\cdot \mid s_t, a_t, Q_L^{\theta}(s_t,a_t))$.
            \STATE Compute policy loss
            \STATE \hspace{1em} $\mathcal{L}_2 = -\log \pi_\theta(a_t^r \mid s_t)$.
            \STATE Update critic and policy parameters
            \STATE \hspace{1em} $\theta \leftarrow \theta - \lambda_1 \nabla_\theta \mathcal{L}_1 - \lambda_2 \nabla_\theta \mathcal{L}_2$.
            \STATE Update target parameters $\bar{\theta} \leftarrow \tau \theta + (1-\tau) \bar{\theta}$.
        \ENDFOR
    \ENDFOR
    \end{algorithmic}
\end{algorithm}

\newpage 

\section{Qualitative Examples}
\label{app:examples}

In this section, we provide qualitative examples from mathematical reasoning, 20Q, and $\tau$-bench that illustrate how the natural language critic supports policy improvement. 
In the mathematical reasoning example, the critic acts like a process reward model (PRM) that pinpoints an incorrect intermediate step, enabling the refinement policy to repair a subtle mistake in a long reasoning chain. In the 20Q example, the critic explicitly encourages exploration of more informative discriminators. In the $\tau$-bench example, the agent partially resolves a request in a way that would violate policy constraints; here, the critic identifies the exact guideline at risk and gives feedback that the LLM can directly use to revise its action. 

\captionsetup[figure]{skip=2pt}
\setlength{\textfloatsep}{6pt}
\setlength{\floatsep}{6pt}
\setlength{\intextsep}{6pt}

\begin{figure}[H]
\centering
\includegraphics[width=\linewidth]{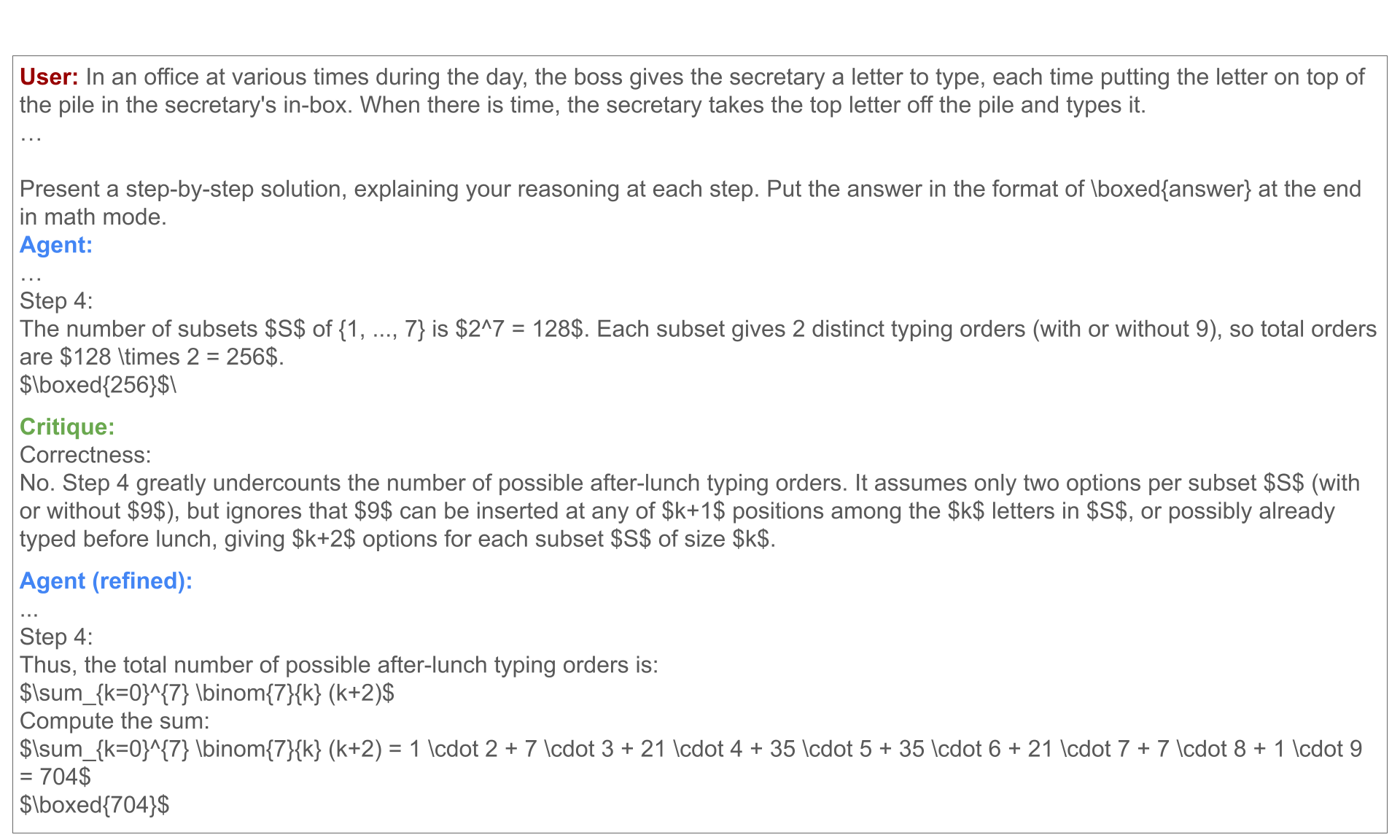}
\caption{\footnotesize Sample timestep on mathematical reasoning. Our learned critic acts as a process reward model, identifying incorrect steps in long reasoning traces. This enables the refinement policy to correct hard-to-find mistakes in just one refinement.}
\label{fig:nlac_example_math}
\vspace{-0.8\baselineskip}
\end{figure}

\begin{figure}[H]
\centering
\includegraphics[width=\linewidth]{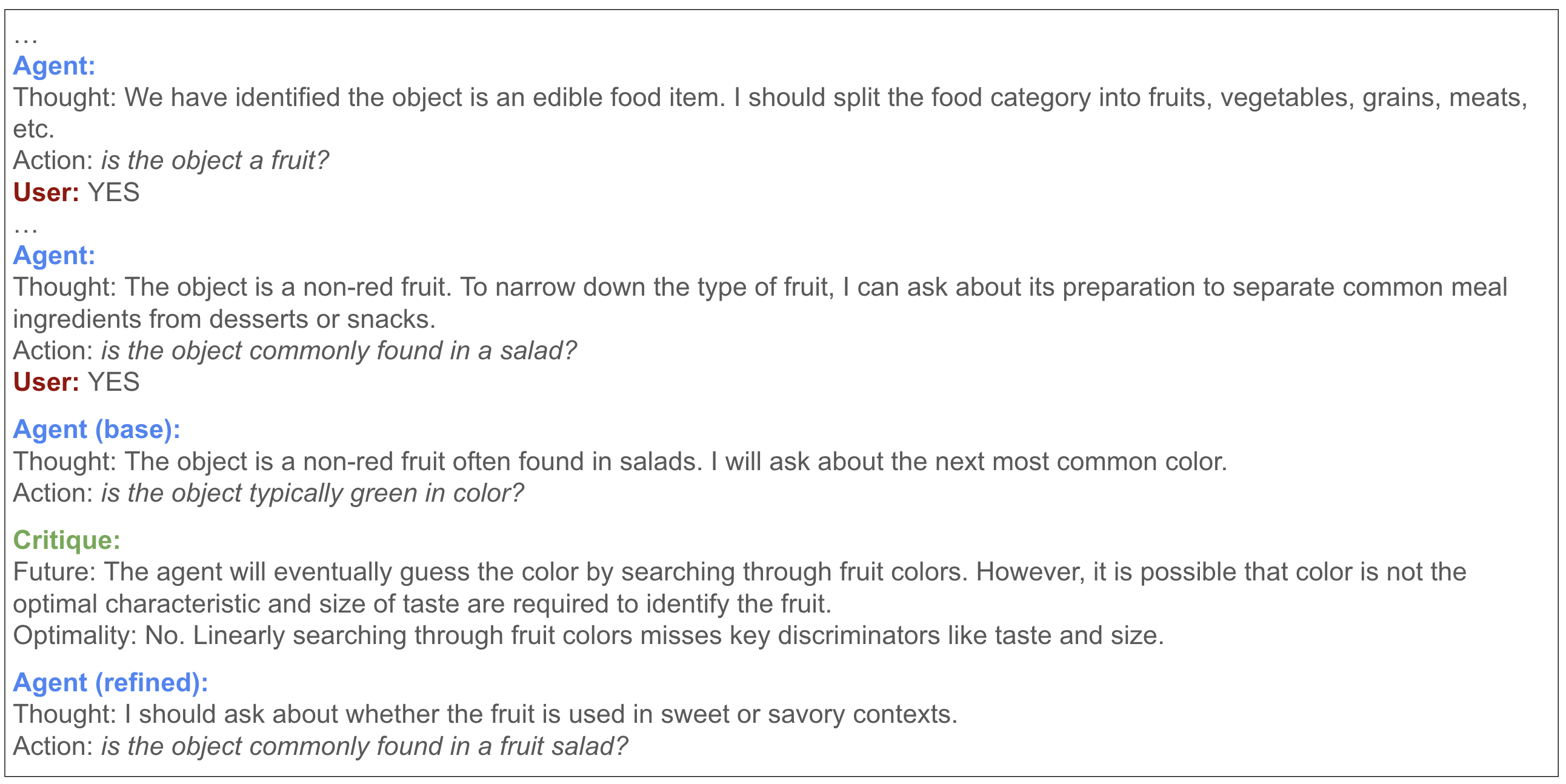}
\caption{\footnotesize Sample timestep on 20Q, where the LLM agent attempts to guess the hidden object ``raisin." The base LLM agent has narrowed down the object to a non-red fruit found in salads, but proceeds to search over the color. However, color is often not the most defining characteristic, so it is more optimal to search over other discriminators such as taste or size.}
\label{fig:nlac_example_20q}
\vspace{-0.8\baselineskip}
\end{figure}

\begin{figure}[H]
\centering
\includegraphics[width=\linewidth]{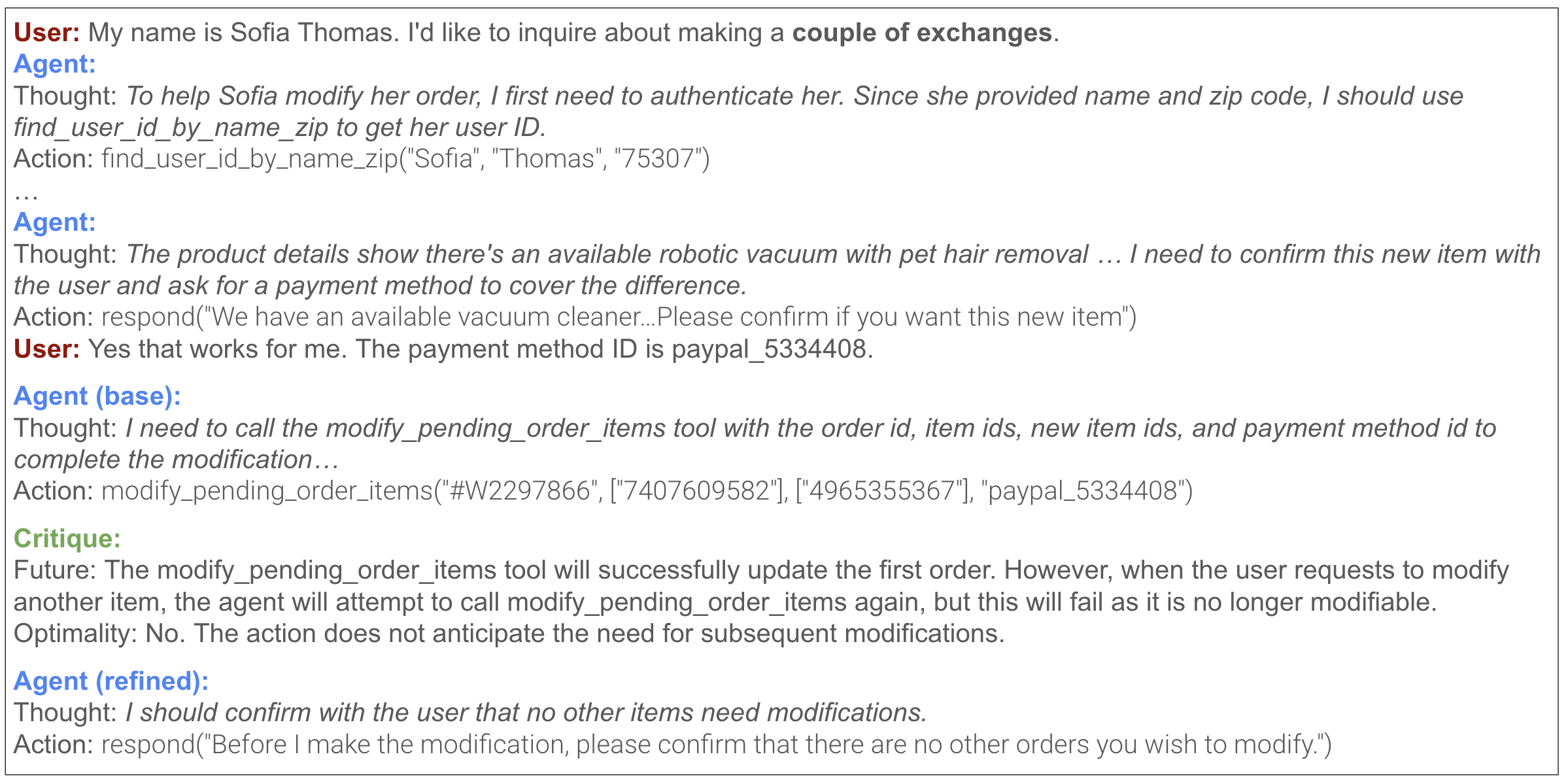}
\caption{\footnotesize Sample timestep on $\tau$-bench where a base LLM agent fails by modifying the database (which can only be done once according to the guidelines) when more exchanges are likely needed. The natural language critic correctly identifies why the action is suboptimal, and explains it in language so that the same LLM can process the critique and correct its action.}
\label{fig:nlac_example_tau}
\vspace{-0.5\baselineskip}
\end{figure}

\newpage

\section{Theoretical Analysis}
\label{app:proofs}

The goal of this section is to show that policy iteration using our proposed NLAC method converges to the optimal policy. 
For traditional actor-critic algorithms, this involves showing (1) convergence of learned Q-values via the Bellman backup, and (2) monotonic improvement of the trained policy. However, such analysis does not apply because the Q-values we learn are textual rather than scalar. 

Instead of analyzing the textual values, we consider the underlying representations that the LLM decodes in order to generate such values. First, we define $\phi(s) \in \mathbb{R}^d$ as features extracted from any state $s$. We assume the following:
\begin{assumption}
The expected reward $r(s, a)$ for any state $s$ and action $a$ can be linearly represented by the features as $r(s, a) = \phi(s) \cdot w$ for some fixed $w \in \mathbb{R}^d$.
\label{ass:reward_linear}
\end{assumption}
Next, we define representations $\Phi^{\pi}(s, a) \in \mathbb{R}^d$ such that we can write the output of our language successor model as $M^\pi(\cdot| s, a) = f_M(\Phi^{\pi}(s, a))$, and similarly $Q_L^\pi(s, a) = f_Q(\Phi^{\pi}(s, a))$, for some functions $f_M, f_Q$ denoting decoding by the LLM.
We make the following assumption about the effect of our proposed language Bellman backup on such representations:
\begin{assumption}
    For any state $s$ and action $a$ , the language Bellman backup satisfies 
    $$\mathcal{B}_L\,M^\pi(\cdot | s, a) = f_M\left(\phi(s) + \gamma \mathbb{E}_{s', a' \sim P^\pi} \, \Phi^{\pi}(s', a')\right)\,.$$
    \label{ass:bellman}
\vspace{-0.3in}
\end{assumption}
While this may initially seem like a strong assumption, note that our language Bellman backup is already instructed to combine the immediate observation with the future description in language space; the assumption only states that the combination also corresponds to a discounted sum in the representation space. Using the above two assumptions, our first main result is the following:
\begin{theorem}
Consider policy evaluation via Equation~\ref{eq:successor_train} and let $Q_L^\pi$ be the natural language critic at convergence. For any state $s$ and action $a$, there exists monotonic mapping $g$ such that $Q^\pi(s, a) = g(Q_L^\pi(s, a))$, where $Q^\pi$ denotes the true scalar Q-function.
\label{thm:evaluation}
\end{theorem}
Our main result makes a precise connection between our learned critic and the true Q-function. The crux of our proof involves showing that our training objective results in a fixed point where $\Phi^{\pi}(s, a)$ are \emph{successor features} of the underlying MDP \citep{barreto2018successorfeaturestransferreinforcement}.

During policy improvement, we update the policy towards a refinement policy $\pi^r$. By definition, the refinement policy is an oracle that for any state $s$ and action $a$, generates $a^r$ such that $Q_L(s, a^r) \geq Q_L(s, a)$. Combining this with the result of Theorem~\ref{thm:evaluation}, we arrive at our second main result:
\begin{theorem}
Repeated application of policy evaluation via Equation~\ref{eq:successor_train} and policy improvement via Equation~\ref{eq:policy_train} from a policy $\pi_0$ converges to policy $\pi^*$ such that $Q^{\pi^*}(s, a) \geq Q^\pi(s, a)$ for any state $s$ and action $a$, and other policy $\pi$. 
\label{thm:iteration}
\end{theorem}
Hence, we are able to show that under the aforementioned assumptions, our approach NLAC can provably find the optimal policy for an underlying MDP.

\subsection{Proof of Theorem~\ref{thm:evaluation}}
The crux of our proof is establishing that the learned representations $\Phi^\pi(s, a)$ converges to the true successor features $\Psi^\pi(s, a)$, and then using properties of successor features to link to the scalar Q-function $Q^\pi$.
Let $\mathcal{R}$ be the metric space of $\mathbb{R}^d$ with the $\ell_{\infty}$ norm.
As shorthand, we also define the iterative operator $\mathcal{T}_{\Phi}^{\pi} \Phi^\pi(s, a) = \phi(s) + \gamma \mathbb{E}_{s', a' \sim P^\pi} \, \Phi^\pi(s', a')$.

First, we verify that $\mathcal{T}_{\Phi}^{\pi}$ is a $\gamma$-contraction mapping on $\mathcal{R}$. For any two functions $\Phi_1, \Phi_2$, we have:
    \begin{align*}
    \| \mathcal{T}_{\Phi}^{\pi} \Phi_1 - \mathcal{T}_{\Phi}^{\pi} \Phi_2 \|_{\infty} &= \sup_{s, a} \| \mathcal{T}_{\Phi}^{\pi} \Phi_1(s, a) - \mathcal{T}_{\Phi}^{\pi} \Phi_2(s, a) \|_{\infty} \\
    &= \sup_{s, a} \left\| \left( \mathbf{\phi}(s) + \gamma \mathbb{E}[\Phi_1(s', a')] \right) - \left( \mathbf{\phi}(s) + \gamma \mathbb{E}[\Phi_2(s', a')] \right) \right\|_{\infty} \\
    &= \gamma \sup_{s, a} \left\| \mathbb{E}_{P^{\pi}}[\Phi_1(s', a') - \Phi_2(s', a')] \right\|_{\infty} \\
    &\leq \gamma \sup_{s, a} \mathbb{E}_{P^{\pi}}\left[ \| \Phi_1(s', a') - \Phi_2(s', a') \|_{\infty} \right] \\
    &\leq \gamma \| \Phi_1 - \Phi_2 \|_{\infty}\,.
    \end{align*}

Since $\mathcal{T}_{\Phi}^{\pi}$ is contraction, the Banach Fixed-Point Theorem with Assumption~\ref{ass:bellman} guarantees that minimization of our training objective in Equation~\ref{eq:successor_train} results in the unique fixed point
$$\Phi^\pi(s, a) = \phi(s) + \gamma \mathbb{E}_{s', a' \sim P^\pi} \, \Phi^\pi(s', a')\,,$$
which are exatly the successor features by definition.

Next, we want to establish a monotonic mapping from representations $\Phi^\pi(s, a)$ to true Q-values $Q^\pi(s, a)$.
The true $Q$-function is defined as $Q^\pi(s_t, a_t) = \mathbb{E}_{\pi} [\sum_{\tau=t}^{\infty} \gamma^{\tau-t} r(s_\tau, a_\tau)]$. Using Assumption~\ref{ass:reward_linear}, and substituting the fixed-point result from earlier, we have:
\begin{align*}
Q^\pi(s, a) &= \mathbb{E}_{\pi} \left[\sum_{\tau=t}^{\infty} \gamma^{\tau-t} (\mathbf{\phi}(s_{\tau}) \cdot w) \right] 
= \left( \mathbb{E}_{\pi} \left[\sum_{\tau=t}^{\infty} \gamma^{\tau-t} \mathbf{\phi}(s_{\tau}) \right] \right) \cdot w 
= \Phi^\pi(s, a) \cdot w\,.
\end{align*}
Since the natural language values at convergence are given by $Q_L^\pi(s, a) = f_Q(\Phi^\pi(s, a))$, the monotonic mapping $g$ is defined by $g(Q_L^\pi(s, a)) = f_Q^{-1}(Q_L^\pi(s, a)) \cdot w$.
\qed

\subsection{Proof of Theorem~\ref{thm:iteration}}

First, we show monotonic policy improvement in each iteration of our algorithm. 
Recall that the refinement policy $\pi^r$ is defined as the oracle that generates $a^r$ such that $Q_L^\pi(s, a^r) \geq Q_L^\pi(s, a)$.
By Theorem~\ref{thm:evaluation}, a monotonic mapping $g$ exists between $Q_L^\pi$ and $Q^\pi$, and ensures that the ordering established in the language space holds in the true scalar space:
$$Q_L^\pi(s, a^r) \geq Q_L^\pi(s, a) \implies Q^\pi(s, a^r) \geq Q^\pi(s, a)\,.$$
Note that the policy improvement objective in Equation~\ref{eq:policy_train} updates $\pi$ to $\pi'$, which is an approximation of the refinement policy $\pi^r$.
By definition of refinement policy, this guarantees that the new policy $\pi'$ is monotonically better than $\pi$ in terms of expected return:
$Q^{\pi'}(s, a) \geq Q^\pi(s, a) \ \text{for all } s, a$. 

Next, we want to show convergence to optimal policy $\pi^*$.
This naturally follows from the fact that in a finite MDP, policy iteration that guarantees monotonic improvement at every step must converge to the unique optimal policy $\pi^*$ in a finite number of iterations.
Hence, at convergence, the policy satisfies $Q^{\pi^*}(s, a) \geq Q^\pi(s, a)$ for all other policy $\pi$.
\qed

\newpage

\section{Implementation Details}
\label{app:imp_details}

In this section, we provide details of implementation of NLAC across the various benchmarks we evaluate. Details include the prompts used to mimic the different components of our algorithm, as well as hyperparameters configured during RL training.

Recall from Section~\ref{sec:implementation} that our algorithm consists of the following novel components: 
\begin{itemize}
\item[(1)] \textbf{language successor model}: probabilistically generates a text prediction of what will happen to policy $\pi$ after taking an action.
\item[(2)] \textbf{language Bellman backup}: uses one-step sample of the immediate next state to also probabilistically generate a target text prediction of the future after taking an action.
\item[(3)] \textbf{language evaluator}: processes textual futures to generate a critique of an action, commenting on optimality and an explanation why by referencing potential future outcomes.
\item[(4)] \textbf{refinement policy}: uses the critique of an action to propose an improved action.
\end{itemize} 
In practice, since number of futures is $k = 1$ in our experiments, we combine the successor model and evaluator into one generation by the \textbf{language critic}.   

\subsection{Language Critic Implementation}
The language critic is implemented by prompting the base LLM with instruction $p_{\text{eval}}(s_t, a_t)$. In the $\tau$-bench benchmark, this is done by appending the following prompt to the history of messages comprising $s_t$ and $a_t$:

\begin{figuredbox}
Evaluate your last action, first predicting one possible future and then comment on whether or not your action was optimal, and if not, how it can be improved. Output should be exactly in the format: \\ 

Future: \\
\textless Predict one possible scenario of what will happen next, up to whether or not you succeed at the long-term task. Be concise and keep to a few sentences at most.\textgreater \\
Optimality: \\
\textless "Yes" or "No". If "No", explain how it can be improved in one sentence using the predicted future to justify your explanation.\textgreater \\ 

Do not generate anything after the evaluation.
\end{figuredbox}

For a single-step task such as mathematical reasoning, the appended prompt is instead:
\begin{figuredbox}
For your attempted solution, please perform the following evaluation and output the result exactly in the format: \\

Correctness: \\
\textless "Yes or No". If "No", identify where any errors occurred. Remember the solution could be incorrect simply because the answer is not formatted correctly with the answer in the format \textbackslash boxed\{answer\}.\textgreater

\end{figuredbox}

\newpage 

\subsection{Language Bellman Backup Implementation}

The language Bellman backup is also implemented by prompting the base LLM with instruction $p_{\text{eval}}(s_t, a_t, s_{t+1})$. This is done by first appending the following prompt to the history of messages comprising $s_t$ and $a_t$ to get a bootstrapped future prediction:
\begin{figuredbox}
The response to your latest action is (could be a tool API output or text utterance from the customer): \\
\{next observation\} \\
From this state, describe one possible scenario of what will happen next, up to whether or not you succeed at the long-term task. Be concise and keep to a few sentences at most.
\end{figuredbox}

Then, the target evaluation is obtained by appending the following prompt afterwards
\begin{figuredbox}
Evaluate your latest action. Remember your output should be in exactly the following format:

Future: \\
\textless Combine the observed response to your latest action with the predicted future from there, up to whether or not you succeed at the long-term task.\textgreater \\
Optimality: \\
\textless "Yes" or "No". If "No", explain how it can be improved in one sentence using the predicted future to justify your explanation.\textgreater \\

Notes:

1. Do not call tools in the evaluation. They will be **ignored**. 

2. If the action is optimal, just say "Yes" after the "Optimality:" tag and do not explain why.

3. Do not generate anything after the evaluation.
\end{figuredbox}

Another important detail when training reasoning models (that output chain-of-thought thinking by default before every generation) is that its chain-of-thought output will reference the next state $s_t$. This makes it an unsuitable training target because it references information not provided to the critic. Hence, we add an additional postprocessing step to generate a \emph{corrected} chain-of-though thinking that removes references to such ground-truth information:
\begin{figuredbox}
In the above evaluation, the chain-of-thought thinking between \textless think\textgreater and \textless\textbackslash think\textgreater tags likely referenced the response to your action and future, or the final score if provided. \\

Fix the chain-of-thought thinking so that it does not refer to those quantities as a reference, but rather infers them. So instead of saying an event will happen in the future, or that the final score is 0, say that you believe it will happen. \\

Your corrected chain-of-thought should be similar to the original in style and prose, but simply remove references to future or the final score as ground-truth information, and instead reason about how you might be able to infer future events from only the observations thus far, up to your latest action. Your output should be in to format:
\textless corrected$\_$think\textgreater Revised chain-of-thought thinking goes here...\textless\textbackslash corrected$\_$think\textgreater \\

It is important that you enclose the corrected chain-of-thought thinking between \textless corrected$\_$think\textgreater and \textless\textbackslash corrected$\_$think\textgreater  tags, as your response will get automatically parsed by a computer. The part after the chain-of-thought thinking should be the evaluation exactly in the format described earlier. \\

There should be exactly one \textless corrected$\_$think\textgreater...\textless\textbackslash corrected$\_$think\textgreater  block in your response. Do not include any \textless think\textgreater  or \textless\textbackslash think\textgreater tags within this block. Do not generate anything after the \textless\textbackslash corrected$\_$think\textgreater tag.
\end{figuredbox}

Then, we extract the corrected chain-of-thought thinking from the output and co-opt the original chain-of-thought-thinking in the target evaluation. 

\subsection{Refinement Policy Implementation}

The refinement policy is implemented by appending an additional prompt after $Q^\pi_L(s_t, a_t)$ that is the output of the language critic:
\begin{figuredbox}
Use the evaluation of the latest action to assess whether the latest action was optimal, and generate a revised action that fixes any problems with the latest action (can simply copy latest action if it is optimal). Output should be exactly in the format: \\

Thought: \\
\textless A single line of reasoning to process the context and inform the decision making. Do not include extra lines.\textgreater  \\
Action: \\
\{"name": \textless Name of action\textgreater, "arguments": \textless Arguments to the action in json format\textgreater \} \\

Note that you are outputting an action that will replace the latest one. Do not output an action that is meant to come afterwards. \\

Do not reference the previous action or its evaluation. \\
\end{figuredbox}

Again, for LLM policies that are reasoning models, we must correct the chain-of-thought thinking that will likely reference the critique (which is not seen by the base policy). We append the following postprocessing prompt afterwards:
\begin{figuredbox}
In the above revised action, the chain-of-thought thinking likely used the previous action and its evaluation to guide your thinking. \\

I want you to fix the chain-of-thought thinking so that it does not use the previous action and its evaluation as reference, but rather infers those quantites. So instead of referring to an action and its evaluation, say that if this action was chosen, then you believe the following evaluation would happen. \\

Your revised chain-of-thought should be similar to the original in style and prose, but motivate the revised action directly from just the last observed tool or customer response, as if the revised action were your first attempt. Your output should be in to format:
\textless corrected$\_$think\textgreater Revised chain-of-thought thinking goes here...\textless\textbackslash corrected$\_$think\textgreater

It is important that you enclose the corrected chain-of-thought thinking between \textless corrected$\_$think\textgreater and \textless\textbackslash corrected$\_$think\textgreater tags, as your response will get automatically parsed by a computer. The part after the chain-of-thought thinking should be the evaluation exactly in the format described earlier.

There should be exactly one \textless corrected$\_$think\textgreater...\textless\textbackslash corrected$\_$think\textgreater block in your response. Do not include any \textless think\textgreater or \textless\textbackslash think\textgreater tags within this block. Do not generate anything after the \textless\textbackslash corrected$\_$think\textgreater tag.

\end{figuredbox}

Like before, we parse the corrected chain-of-thought thinking and replace the original thinking in the output of the refinement policy.

\newpage 

\subsection{Training Details}
\label{app:training_details}
Our fine-tuning baselines were implemented using the Volcano Engine Reinforcement Learning (verl) library~\citep{sheng2024hybridflow}. We train on $8$ H20 GPU nodes, resulting in $64$ GPUs total, for a total of $30,720$ gradient steps. Training took $<48$ hours for each benchmark. We used the following hyperparameter configuration for each benchmark, after some minimal amount of tuning:
\begin{table}[H]
\centering
\begin{tabular}{lrr}
\toprule
Hyperparameter & \multicolumn{2}{r}{Setting } \\
\midrule
\midrule
Maximum prompt length && 8192 \\
Maximum response length && 24576 \\
Batch size && 1024 \\
Number of iterations && 30 \\
Target network update $\tau$ && 0.005  \\
Prioritized replay buffer $\alpha$ && 0.1 \\
Optimizer && AdamW \\
Learning rate && 5e-6 \\
\bottomrule
\end{tabular}   
\end{table}

In practice, we found it helpful to implement a prioritized replay buffer weighted by $\mathcal{L}_1(s_t, a_t, s_{t+1})$ with sampling parameter $\alpha$ \citep{schaul2016prioritizedexperiencereplay}. 
This is because in many tasks, though a base LLM policy may achieve low reward in a large proportion of rollouts, many actions in these unsuccessful rollouts are still optimal. 
Therefore, to improve learning efficiency, we prioritize training on samples where the agent is likely to take a suboptimal action, using critic loss as a proxy for the likelihood.

\subsection{Training the Refinement Policy}
\label{app:refinement_training}
Currently, our method relies on base reasoning capabilities for the refinement policy to generate $a_t^r \sim \pi^r_\theta(\cdot | s_t, a_t, Q_L^\theta(s_t, a_t))$ such that $Q_L^\theta(s_t, a_t^r) \geq Q_L^\theta(s_t, a_t)$. In situations where pretrained LLMs cannot refine actions through in-context learning, we describe how to explicitly train the refinement policy. 

To train the refinement policy, we adopt a similar approach as \citet{kumar2024traininglanguagemodelsselfcorrect} did to train policies to self-correct, but generalized to multi-step MDPs. Namely, for some on-policy sample $s_t, a_t, a_t^r$ from the refinement policy, we additionally train on the loss:
\begin{align}
\mathcal{L}_r(s_t, a_t, a_t^r)  =  -\log \pi_\theta(a_t^r \mid s_t) \left(A^{\pi_{\theta}^r}(s_t, a_t^r) + \alpha\left(A^{\pi_{\theta}^r}(s_t, a_t^r) - A^{\pi_{\theta}}(s_t, a_t)\right)\right)\,.    
\end{align}
Here, $A$ is the estimated advantage function, which is either learned (as in PPO), or obtained from averaging (as in GRPO), using Monte-Carlo rewards. We also include a bonus that is the improvement in advantage of refined $a_t^r$ over base $a_t$, where $\alpha > 0$ is a tunable parameter.
Our findings indicated that this approach resulted in only marginal performance gains compared to the simpler, implicit distillation method already employed by NLAC. Furthermore, the explicit RL objective substantially increased training cost, requiring on-policy samples. Therefore, in the experiments we consider, we do not employ such training. However, in more complex tasks, this may be necessary.

\subsection{Mitigating Catastrophic Forgetting}
\label{app:lwf}

Several components of our method, such as the language Bellman backup and refinement policy, are never explicitly trained but merely prompted to behave according to instruction. However, a key challenge that arises is as our LLM model is trained, it becomes increasingly less competent at following such instructions. We eventually observe \emph{catastrophic forgetting} across all our training runs, resulting in the score during training collapsing to $0$.

In our experiments, we choose to stop training before catastrophic forgetting became a noticeable issue. 
However, we also considered methods to minimize forgetting, notably by incorporating learning without forgetting into our training objective ~\citep{li2016learningwithoutforgetting}. 
The way we do so is by introducing auxiliary loss functions that penalizes divergence from the initial, pre-trained LLM.
We introduce two additional losses for the language Bellman backup and refinement policy, respectively:
\begin{align}
    \mathcal{L}_{\mathrm{lwf}, 1}(s_t, a_t, r_t, s_{t+1}) &= D_{\mathsf{KL}}\left(\mathcal{B}_L M_{\theta}(\cdot \mid s_t, a_t) \ || \ \mathcal{B}_L M_{\mathrm{init}}(\cdot \mid s_t, a_t) \right) \\
    \mathcal{L}_{\mathrm{lwf}, 2}(s_t, a_t, Q_L(s_t, a_t)) &= 
    D_{\mathsf{KL}}\left(\pi^r_{\theta}(\cdot \mid s_t, a_t, Q_L(s_t, a_t)) \ || \ \pi^r_{\mathrm{init}}(\cdot \mid s_t, a_t, Q_L(s_t, a_t)) \right) \nonumber
\end{align}

Ultimately, we found that such objectives were able to prevent collapse, allowing our LLM model to be trained for more iterations. However, training for more iterations did not prove significantly advantageous in improving final performance. Therefore, we did not include such objectives in our default algorithm.

\newpage
\section*{NeurIPS Paper Checklist}
\begin{enumerate}[leftmargin=*,label=\arabic*.]
\item \textbf{Claims}\
      \answerYes{} The paper states its main claims in the abstract and introduction, and revisits them in the discussion.

\item \textbf{Limitations}\
      \answerYes{} Limitations are discussed in the main paper, especially in the discussion section.

\item \textbf{Theory Assumptions and Proofs}\
      \answerYes{} Assumptions are stated in the theoretical analysis appendix, and proofs are provided there as well.

\item \textbf{Experimental Reproducibility}\
      \answerYes{} The paper describes the algorithm, provides appendix pseudocode, and includes implementation/training details in the appendix.

\item \textbf{Open Access to Data and Code}\
      \answerNo{} We do not include a public code or data release in this submission. Our implementation leverages open-source RLVR pipelines, and we plan to release the training code for our newly proposed algorithm in a public release of the paper. 

\item \textbf{Experimental Setting and Details}\
      \answerYes{} The paper specifies benchmarks, models, baselines, training budgets, and key hyperparameters, with additional details in the appendix.

\item \textbf{Statistical Significance}\
      \answerYes{} Reported results are averaged across three independent runs where applicable, and the learning-curve figure includes error bars.

\item \textbf{Compute Resources}\
      \answerYes{} The appendix documents the hardware setup, training duration, and overall training budget.

\item \textbf{Code of Ethics}\
      \answerYes{} We affirm that we have read and will adhere to the NeurIPS Code of Ethics.

\item \textbf{Broader Impacts}\
      \answerNo{} We do not include a dedicated broader impacts section. Our work proposes a new method of training LLM agents with a mix of theoretical and empirical evidence. We do not anticipate any concerns on societal impact beyond any other research in developing better RL algorithms.

\item \textbf{Safeguards and Risk Mitigation}\
      \answerNA{} This submission does not release a new model, dataset, or system artifact for deployment.

\item \textbf{Licenses for Existing Assets}\
      \answerNo{} The paper uses existing benchmarks and models, but it does not yet include a dedicated summary of licenses and terms of use.

\item \textbf{New Assets}\
      \answerNA{} The submission does not introduce a new dataset, model, or other redistributed asset.

\item \textbf{Human Subjects or Crowdsourcing}\
      \answerNA{} The experiments do not involve crowdsourcing or research with human subjects.
\end{enumerate}

\end{document}